# X-CAR: An Experimental Vehicle Platform for Connected Autonomy Research Powered by CARMA<sup>SM</sup>

Goodarz Mehr, Prasenjit Ghorai, Ce Zhang, Anshul Nayak, Darshit Patel, Shathushan Sivashangaran, and Azim Eskandarian, *Senior Member, IEEE*

*Abstract*—Autonomous vehicles promise a future with a safer, cleaner, more efficient, and more reliable transportation system. However, the current approach to autonomy has focused on building small, disparate intelligences that are closed off to the rest of the world. Vehicle connectivity has been proposed as a solution, relying on a vision of the future where a mix of connected autonomous and human-driven vehicles populate the road. Developed by the U.S. Department of Transportation Federal Highway Administration as a reusable, extensible platform for controlling connected autonomous vehicles, the CARMA Platform<sup>SM</sup> is one of the technologies enabling this connected future. Nevertheless, the adoption of the CARMA Platform<sup>SM</sup> has been slow, with a contributing factor being the limited, expensive, and relatively old vehicle configurations that are officially supported. To alleviate this problem, we propose X-CAR (eXperimental vehicle platform for Connected Autonomy Research). By implementing the CARMA Platform<sup>SM</sup> on more affordable, high quality hardware, X-CAR aims to increase the versatility of the CARMA Platform<sup>SM</sup> and facilitate its adoption for research and development of connected driving automation.

*Index Terms*—Connected mobility, connected autonomous vehicles, intelligent transportation systems, experimental vehicle platform, CARMA<sup>SM</sup>, tutorial

## I. INTRODUCTION

AUTONOMOUS vehicles (AVs) promise a future with a safer, cleaner, more efficient, and more reliable transportation system [1], [2]. At high penetration rates they can reduce traffic accident fatalities and pedestrian collisions, lower fuel consumption, utilize efficient routes, and reduce traffic congestion [3]–[5]. However, the current approach to autonomy has focused on building small, disparate intelligences that are closed off to the rest of the world. In this approach, even if several autonomous vehicles are traveling in the same environment at the same time, they each have to carry expensive sensing, navigation, and processing hardware and still, lacking coordination with other road users, they may get into accidents.

Vehicle connectivity has been proposed as a solution, relying on a vision of the future where a mix of connected

Corresponding author: Goodarz Mehr

The authors are with the Autonomous Systems and Intelligent Machines (ASIM) Laboratory, Virginia Tech, Blacksburg, VA 24061, USA. (email: goodarzm@vt.edu; prasenjitg@vt.edu; zce@vt.edu; anshulnayak@vt.edu; darsh2198@vt.edu; shathushansiva@vt.edu; eskandarian@vt.edu).

autonomous and human-driven vehicles populate the road. Connectivity allows vehicles to perceive beyond the field of view (FoV) of their sensors, coordinate with other vehicles, and negotiate with other road users [1].

The CARMA<sup>SM</sup> (Cooperative Automation Research Mobility Applications) program at the U.S. Department of Transportation Federal Highway Administration (FHWA)a is one of the leading forces behind cooperative driving autonomy (CDA) research, exploring the application of CDA to traffic, reliability, and freight scenarios [6]. CARMA<sup>SM</sup> products provide the necessary software for conducting CDA research and testing, and include CARMA Cloud<sup>SM</sup>, CARMA Platform<sup>SM</sup>, CARMA Messenger, and CARMA Streets [7]. All four products are open-source and work together with the FHWA V2X Hub, a separate multi-modal system enabling networked, wireless communication between connected autonomous vehicles (CAVs), infrastructure modules, and personal communication devices [8].

The CARMA Platform<sup>SM</sup> is a reusable, extensible platform for controlling SAE level 3+ connected autonomous vehicles (CAVs). It is written in C++ and runs in a Robot Operating System (ROS) environment on Ubuntu. It's latest version is CARMA3, which provides a rich, generic API for various sensors and actuators, as well as for third-party plugins that implement guidance algorithms to plan vehicle trajectories [9], [10]. The adoption of CARMA3 by the research community, however, has been slow. One contributing factor may be the fact that officially-supported vehicle configurations are limited, expensive, and use hardware that is several years old. To alleviate this problem, we developed X-CAR (eXperimental vehicle platform for Connected Autonomy Research). By implementing CARMA3 on a wider set of more affordable, high quality hardware, X-CAR aims to increase the versatility of CARMA3 and facilitate its adoption for research and development (R&D) of CDA. We used a 2017 Ford Fusion SE Hybrid for X-CAR development which can be seen in Fig. 1, and documented our work in the X-CAR Reference Manual [11]. It should be noted that the use of any particular vehicle, sensor, or hardware in this paper is not a commercial endorsement of the manufacturer. For most items alternatives are available and can be used without sacrificing the integrity of the system. Nevertheless, hardware-specific information is provided here for the sake of accuracy and to facilitate the reproducibility of the platform and results.

The remainder of this document is organized as follows: Section II reviews past experimental AV and CAV platforms and provides some introductory information on topics such as ROS and Docker, Section III and Section IV examine the sensors and actuators that comprise X-CAR, Section V looks at X-CAR's V2X setup, Section VI discusses X-CAR's computation and data processing system, Section VII provides an overview of CARMA3 and its subsystems, Section VIII provides information on the type of maps and routes used by CARMA3 for localization and guidance, Section IX evaluates X-CAR's real-world performance, and finally Section X concludes this paper. A guide to installing CARMA3 on X-CAR, configuring the software, and operating X-CAR is provided in Appendix A.

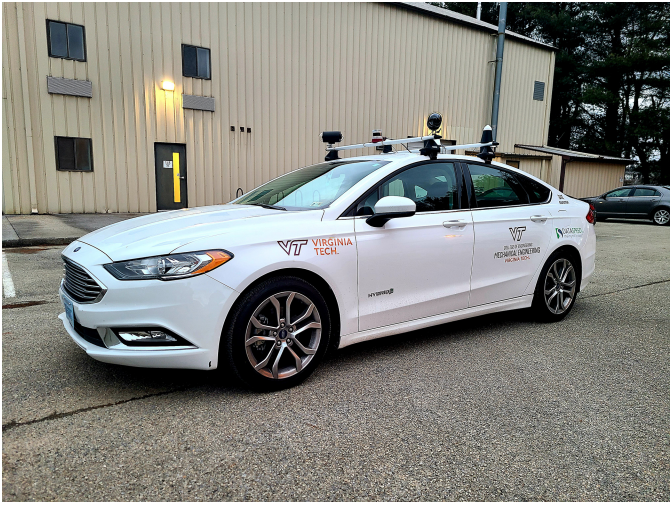

Fig. 1. 2017 Ford Fusion SE Hybrid used for X-CAR development.

TABLE I
PERCEPTION SENSOR CONFIGURATION OF DIFFERENT AV PLATFORMS

| Platform | Lidar | Radar | Camera |
|---|---|---|---|
| BRAiVE [15] | 5 | 0 | 10 |
| Deeva [16] | 4 | 0 | 26 |
| RobotCar [19] | 3 | 0 | 6 |
| Tesla FSD [20] | 0 | 0 | 8 |
| Uber Volvo XC90 [21] | 1 | 10 | 7 |
| Uber Ford Fusion [21] | 7 | 7 | 20 |
| Aptive Pro [22] | 1 | 6 | 8 |
| Bertha [23] | 0 | 8 | 3 |
| GM Cruise [24] | 5 | 21 | 16 |
| Nvidia DRIVE Hyperion 9 [25] | 3 | 9 | 14 |
| Waymo Driver 5th Generation [26], [27] | 5 | 6 | 29 |
| Argo Autonomy Platform [28] | 6 | 11 | 12 |
| CARMA Kit [29] | 1 | 5 | 2 |
| X-CAR | 1 | 1 | 7 |

## II. BACKGROUND

In this section, we first review some past experimental AV and CAV platforms and then briefly introduce ROS, Docker, and Autoware®.

### A. Past Experimental AV and CAV Platforms

CAV platforms are AV platforms with added Wireless and cellular communication modules. Some of the earliest experimental AV platforms were vehicles such as Carnegie Mellon University's Boss, Stanford University's Junior, and Virginia Tech's Odin that competed in the 2007 DARPA Urban Challenge [12]–[14]. These early platforms incorporated multiple cameras, radars, and ultrasonic sensors; 2D laser range finders (a precursor to modern lidars); a GPS/IMU module, vehicle actuators; computation and data processing units; and in some cases a then-new Velodyne HDL-64 lidar.

Subsequent platforms followed a similar setup. The BRAiVE AV platform consisted of 10 cameras, 5 lidars, a GPS/IMU module, and vehicle actuators [15]. Similarly, the Deeva AV platform was equipped with 26 cameras divided into 13 stereo pairs, 4 lidars, a GPS/IMU module, and vehicle actuators [16]. AV platforms developed by industry follow a similar pattern and, with the exception of Tesla Full Self-Driving (FSD) that does not use lidar or radar, generally consist of multiple cameras, lidars, radars, and ultrasonic sensors for perception of the surrounding environment; a GNSS/INS module for localization; vehicle actuators; and a computation and data processing unit. A summary of the perception sensor configuration of these platforms can be seen in Table I [17].

The CARMA^SM development team uses a CAV platform configured by AutonomouStuff and called CARMA Kit [18]. CARMA Kit includes a Velodyne Ultra Puck lidar, multiple Allied Vision Mako G-319C cameras, Aptive ESR 2.5 and SRR2 radars, a NovAtel PwrPak7D-E2 GNSS/INS module with 2 Mobile Mark SMW-303 antennas, 2 Cohda Wireless MK5 OBU dedicated short-range communication (DSRC) units, an AStuff Spectra processing unit, and a drive-by-wire (DbW) kit. While CARMA Kit can provide reasonable performance, a six-figure price tag makes it prohibitively expensive. Therefore, X-CAR relies on more affordable components that match or exceed the quality of those used in CARMA Kit to increase the versatility of CARMA3 and facilitate its adoption for R&D of CDA.

### B. ROS

ROS is an open-source robotics middleware suite that is not an operating system (OS) in the traditional sense of process scheduling and management, but rather a collection of frameworks for robot software development [30], [31]. It provides services such as communication between processes, hardware abstraction, low-level device control, and package management for a heterogeneous computer cluster. Nodes structured in a graph architecture conduct processing and may receive, publish, and manipulate perception, control, and other messages. Though not a real-time OS (RTOS), it is possible to integrate ROS with real-time code.

The original ROS, or ROS 1, began in 2007 and its last long-term support (LTS) distribution release was ROS Noetic Ninjemys, primarily targeting Ubuntu 20.04 (Focal Fossa) release and supported until May 2025 [32]. The new ROS, or ROS 2, is a major revision of the ROS application-program interface (API) that will take advantage of modern technologies and libraries for core ROS functionality and add support for embedded hardware and real-time code. CARMA3 was originally written using ROS 1, but is in the process of migrating to ROS 2, at which point it will become CARMA4. For the time being, X-CAR drivers are written using ROS 1.

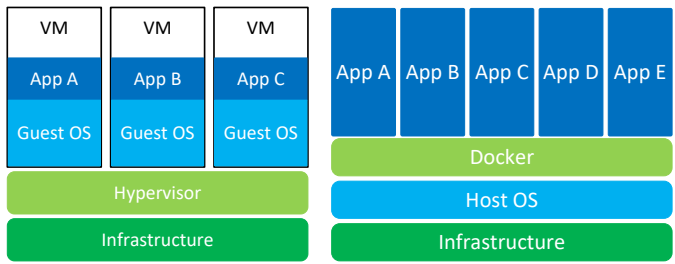

Fig. 2. Comparison between virtualization (left) and containerization (right).

### C. Docker

Docker is an open platform that uses OS-level virtualization to package software in containers. Containers are isolated from one another (and the host OS to some extent) and bundle their own configuration files, libraries, and software. Well-defined channels allow containers to communicate with each other. The software that hosts the containers is called the Docker Engine [33].

Containerization using Docker differs from virtualization. VirtualBox and VMWare are virtualization software that create virtual machines that are isolated at the hardware level and each have their own OS [34], [35]. In contrast, Docker isolates software at the OS level. With Docker, the user can run multiple applications (containers) on the same host OS, sharing underlying hardware resources (CPU, RAM, etc.) [36]. Fig. 2 illustrates the difference between these two approaches.

A Docker image can be thought of as a starting point for building containers, like a template. A Docker image can be built using a `Dockerfile`, which is a set of instructions that can, for example, define the parent image, install a series of packages using a package manager like `apt`, build some packages from source using `catkin` or `colcon`, and determine what commands are executed each time a container is created from the image.

A Docker container is by design isolated from the rest of the OS, hence it has its own file system and by default cannot access any I/O devices (such as USB ports) or any network the OS is connected to. Because of this, the user has to specify the volumes (folders) and I/O devices from the OS it has access to and how it handles networking. While these can be specified as options to the `docker run` command, it is often easier to specify them in a `.yml` Docker Compose file.

### D. Autoware®

Autoware® is a ROS-based open-source software for autonomous vehicles. It supports camera, lidar, radar, and GNSS/INS as primary sensors, and its general structure is depicted in Fig. 3 [37]–[39].

Autoware® has modules for localization, detection, tracking and prediction, planning, and control. The localization module uses point cloud map data and the Normal Distribution

Transform (NDT) algorithm to find a match for successive lidar scans within the point cloud map. The resulting pose is further enhanced by a Kalman filter, using odometry information obtained from CAN and GNSS/INS data. The detection module uses deep learning algorithms to detect nearby objects from the fused camera, radar, and lidar data. The result is fed into a Kalman filter by the tracking and prediction module to better understand the behavior of surrounding dynamic objects. The planning module incorporates probabilistic robotics, rule-based systems, and deep learning in some areas to chart a route, a maneuver plan, and a trajectory for the vehicle. Finally, the control module provides the desired twist of linear and angular velocity to the low-level vehicle controller. It falls into both the Autoware®-side stack (Model Predictive Control (MPC) and Pure Pursuit) and the vehicle-side interface.

## III. SENSING

The X-CAR platform consists of 7 cameras, a mechanical lidar, a long-range radar, a dual-antenna GNSS/INS module, a single-antenna DSRC module, a DbW kit, a processing unit, and a power distribution system, with plans to add a cellular communication module in the future. The planned location and FoV of the sensing and communication equipment is illustrated in Fig. 4 and the planned X-CAR wiring diagram is shown in Fig. 5.

### A. Camera

Cameras are passive sensors in the sense that they capture the light emitted from the surrounding environment and use it to create a monochrome or color image. The resulting image can be used for road (surface and lane line) detection, regulatory element (traffic sign and traffic light) detection, and object detection and tracking [40], [41].

Cameras come in many different forms and use a variety of image sensors and interfaces (GigE Vision [42], USB3 Vision [43], GMSL2 [44], FPD-Link [45], etc.). Some cameras are also available as a stereo pair. For X-CAR, we limited our search to automotive and machine vision cameras, and prioritized the following qualities in our search: dynamic range, resolution, frame rate, IP rating, and cost.

Our initial setup comprised of dual Leopard Imaging LI-IMX390-GMSL2-200H cameras along with a Connect Tech deserializer board mounted on an Nvidia Jetson AGX Xavier Developer Kit, which acted as a centralized Image Signal Processor (ISP) [46]–[49]. The cameras use the GMSL2 interface, have a 200 degree horizontal FoV, and use Sony's IMX390 image sensor which has an advertised dynamic range of 120 dB, a resolution of 1920×1080, and a frame rate of 30 frames per second (FPS) at the maximum resolution [50]. While these cameras were relatively inexpensive, had a high throughput, and worked well in our experiments in the past few years, after encountering issues with driver behavior, color accuracy, dynamic range, and vignetting caused by the large FoV, we decided to upgrade our setup to include more cameras with smaller horizontal FoV.

To that end, our new setup is comprised of 7 Lucid Vision Labs TRI054S-CC cameras [51]. They use the GigE Vision

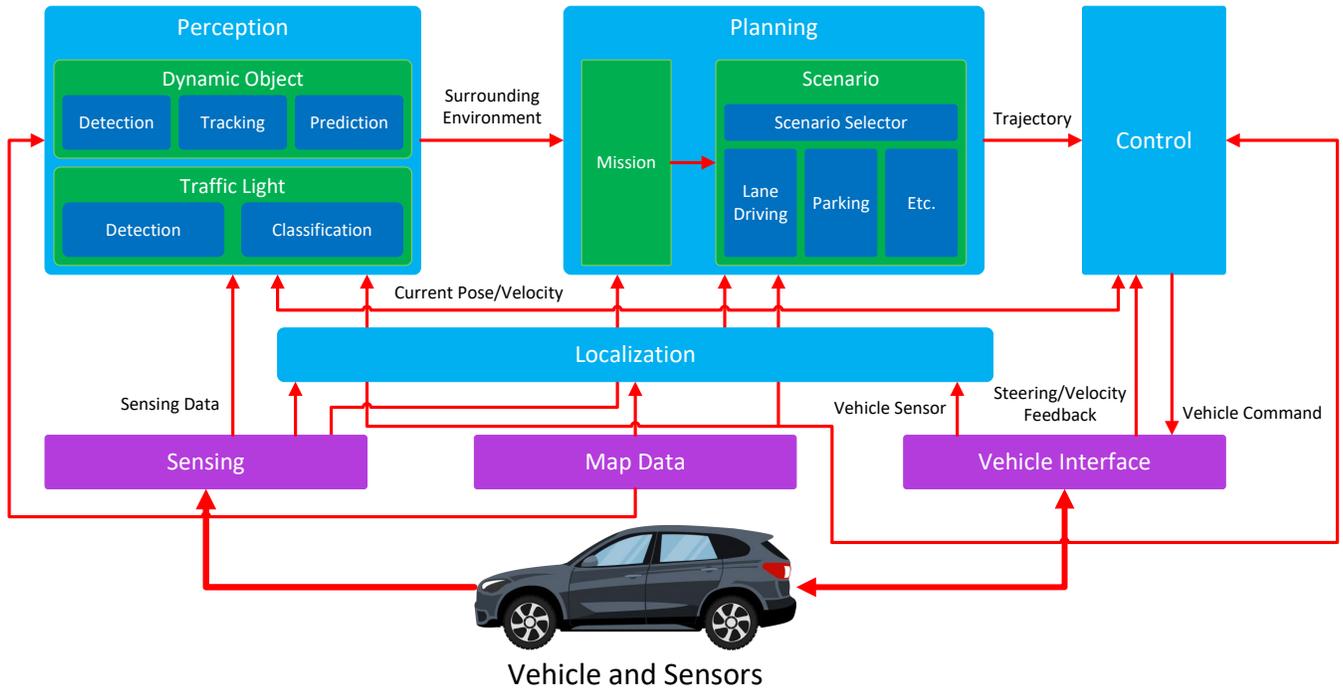

Fig. 3. Overview of Autoware® [37].

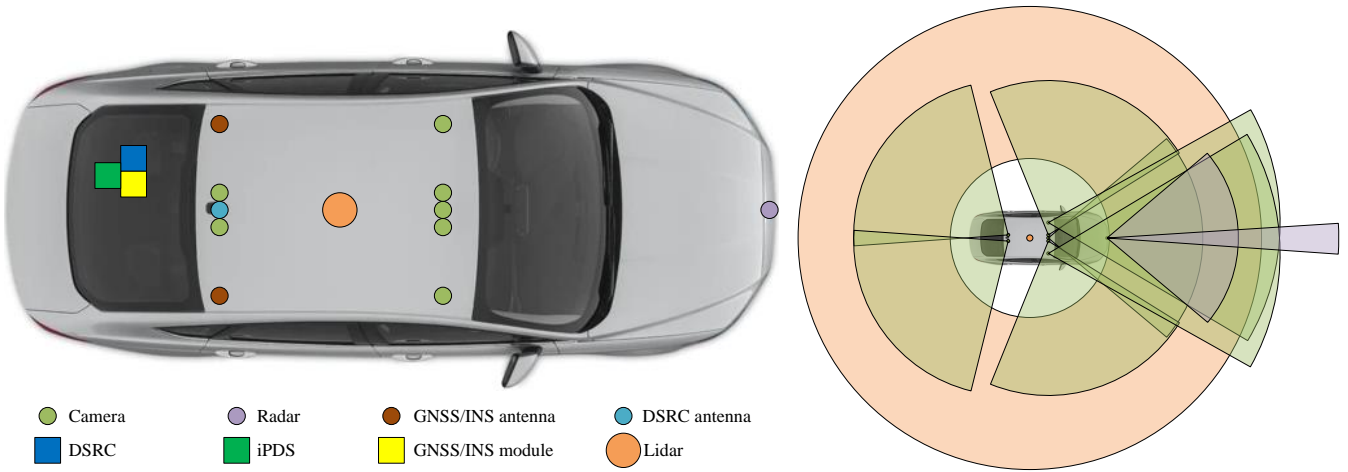

Fig. 4. Planned X-CAR sensing and communication equipment location (left) and FoV (right). Figures are not up to scale.

interface, with 5 of them using the Universe compact C-mount 6mm f/2.0 lens with an 80 degree horizontal FoV (to cover the vehicle surrounding in a pentagon-like pattern) and the other 2 using the Universe compact C-mount 8mm f/2.0 lens with a 60 degree horizontal FoV for forward-looking stereo use [52], [53]. They use Sony's newer IMX490 image sensor which has LED flicker mitigation (LFM) technology, an advertised dynamic range of 120 dB, a resolution of 2880×1860, and a frame rate of 20.8 FPS at the maximum resolution [54]. The cameras have an IP67 rating but the accompanying lenses need to be placed in lens tubes to achieve IP67 rating. Accounting for the accessories (lens, lens tube, cables, mount), each camera costs ~$1,000. The integration of our new camera setup has been delayed due to Covid-19-induced disruptions, which has also delayed our sensor synchronization and processing unit installation plans.

The camera setup consists of the components listed in Table II. After attaching each camera to its respective lens and lens tube, it should be positioned on top of the car as shown in Fig. 4. The cameras should be connected to a Gigabit network switch via the M12 to RJ45 cables. The network switch has to be capable of Power over Ethernet (PoE) and have at least one 10Gbps port for connection to the processing unit. Each camera should also be connected to the GNSS/INS module via the M8 GPIO cable for synchronization. Finally, each camera needs to be calibrated for distortion and color accuracy. Further information in this regard can be found in the Triton 5.4 MP Technical Reference Manual [55].

The Lucid Vision Labs Camera Driver for CARMA GitHub repository contains the X-CAR driver developed for Lucid Vision Labs cameras [56]. Instructions are provided for downloading the driver code, building a Docker image of the driver, and setting up that image to launch with CARMA3. In addition to installation instructions, that repository also contains information on the driver's ROS API. The driver is forked from Lucid Vision Labs' ROS driver with two modifications [57]. First, the X-CAR driver broadcasts a discovery message indicating its current status (operational, degraded, fault, or off) to CARMA3. Second, it publishes camera images in the format expected by CARMA3.

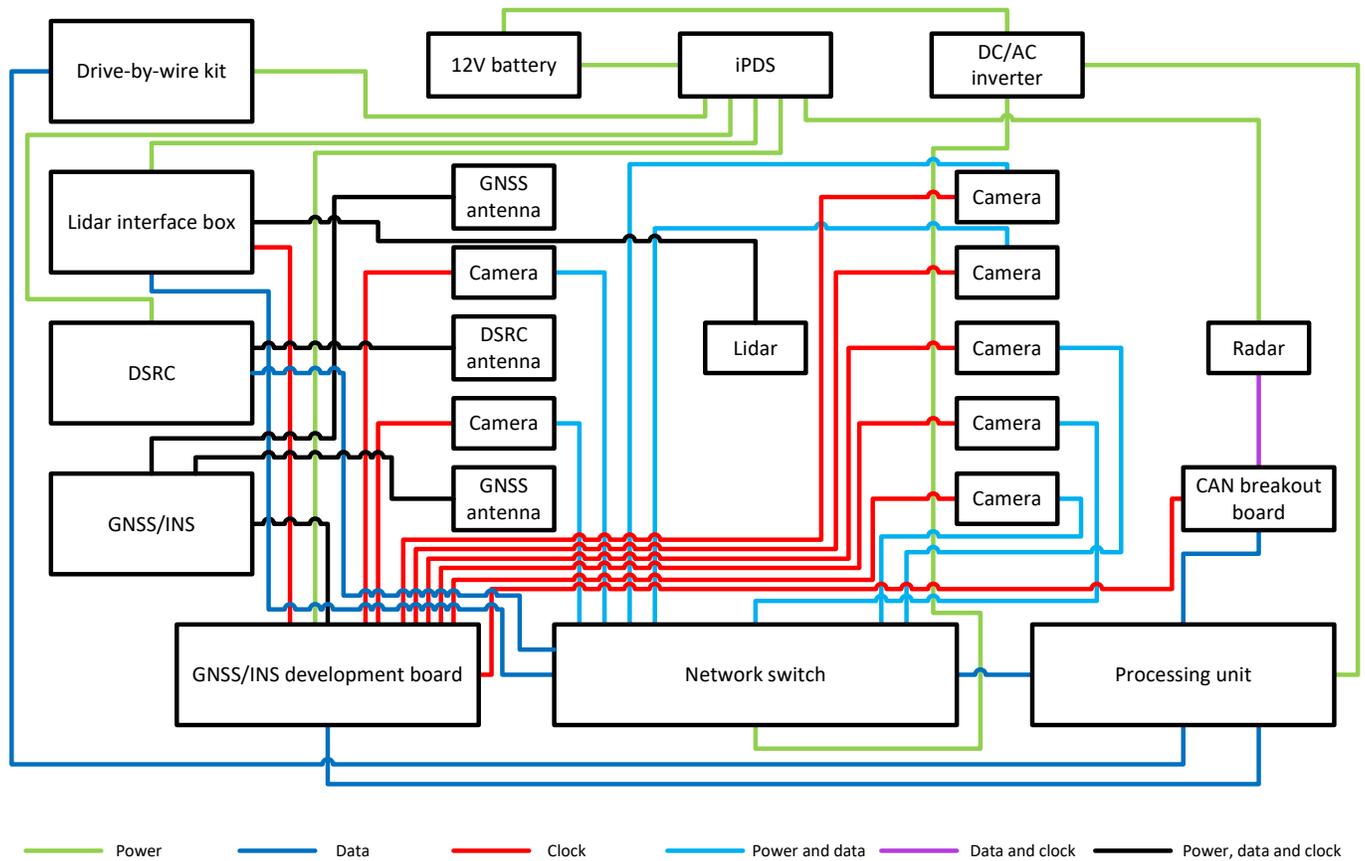

Fig. 5. Planned X-CAR wiring diagram.

**TABLE II**
**Camera Setup Components**

| Quantity | Item |
|---|---|
| 7 | Lucid Vision Labs TRI054S-CC camera |
| 5 | Universe BL060C lens |
| 2 | Universe BL080C lens |
| 7 | Lucid Vision Labs IPTC-D355L399 IP67 lens tube |
| 7 | Lucid Vision Labs CAB-MR-5M-A M12 to RJ45 IP67 Cat6a cable |
| 7 | Lucid Vision Labs GPIO-M8-5M M8 GPIO cable |
| 7 | Lucid Vision Labs PHX-TR tripod adapter |

### B. Lidar

A Lidar perceives its surrounding environment by sending out laser beams and measuring the time it takes for the reflected light to return to the receiver. Using this method, a lidar can create a detailed point cloud map of its surroundings that can be used to detect different objects and surfaces. The main obstacle to the use of lidars is their extremely high cost.

A wide variety of lidars are available on the market. Based on their scanning mechanism, lidars can be divided into mechanical (where a spindle spins to give a 360 degree view), solid state (where the field of view is fixed and there are no moving parts, and either MEMS or optical phase arrays are used to steer the beam), and flash (where a light is flashed over a large field of view) lidars. After careful evaluation, we selected the Ouster OS1-64U lidar for X-CAR, with a range of 120 meters and 64 beams arranged across a 45 degree vertical FoV with uniform spacing [58]. It provides a detailed, 360 degree view of the surrounding environment with a precision of $\pm 5$ cm, has an IP68 rating, and is relatively affordable at about a quarter of the price of a Velodyne Ultra Puck used in the CARMA Kit. It comes with a 2-year warranty.

The lidar setup consists of the components listed in Table III. The sensor should be mounted on a platform above the car in a way that no part of the car's body intersects with any of the laser beams, as shown in Fig. 6. The platform should be level, the sensor should be laterally centered (be equidistant from the right and left sides of the car), and the connector port should directly face the back of the car. This is because in the sensor's internal coordinate frame the connector port is located in the $-x$ direction (with the $z$ direction being vertically upward) [59]. We recommend against taking the sensor's baseplate off, since it also acts as a heat sink and taking it off can cause the sensor to overheat during operation and eventually die.

TABLE III
LIDAR SETUP COMPONENTS

| Quantity | Item |
|----------|------|
| 1 | Ouster OS1-64U lidar |
| 1 | Ouster sensor to Interface Box cable |
| 1 | Ouster Interface Box and 24V power supply |
| 1 | RJ45 cable |

The Interface Box should be installed in the vehicle's trunk and connected to a 24V DC power supply, as shown in Fig. 7. It should be connected to the sensor using the provided cable, to the network switch (or the processing unit) using the provided RJ45 cable, and to the GNSS/INS module for synchronization. Once connected, visiting `http://os-[serial number].local/` on a web browser grants access to the sensor's home page, where the user can install firmware upgrades, see diagnostic information, and change configuration parameters. The sensor is pre-calibrated and does not require any calibration. For further information, consult [59]–[62].

The Ouster Lidar Driver for CARMA GitHub repository contains the X-CAR driver developed for Ouster lidars [63]. Instructions are provided for downloading the driver code, building a Docker image of the driver, and setting up that image to launch with CARMA3. In addition to installation instructions, the repository also contains information on the driver's ROS API. The driver is forked from Ouster's ROS driver with two modifications [64]. First, the X-CAR driver broadcasts a discovery message indicating its current status (operational, degraded, fault, or off) to CARMA3. Second, it uses Unix time as timestamp for published point cloud messages instead of the lidar's internal clock that starts from zero each time the sensor is turned on, in line with CARMA3.

### C. Radar

Radars use radio waves to determine the relative distance, angle, and velocity of the surrounding objects. They transmit electromagnetic waves in the radio or microwave spectrum and process the received reflection to determine the characteristics of surrounding objects. Radars developed for automotive applications generally use frequency-modulated continuous-wave (FMCW) technology [65].

Automotive radars are generally categorized as either short-range radars (SRRs) or long-range radars (LRRs). SRRs operate in the 24 GHz band, have a wide FoV, and a maximum range of about 100 m. In contrast, LRRs operate in the 77 GHz band, have a narrow far-field FoV and a wider near-field FoV, and their maximum range for detecting certain objects can reach even 1 km.

A variety of automotive radars are available on the market, but for low-volume research applications the options are more limited. After careful consideration of available options, we selected the Continental ARS 408-21 LRR [66]. It has a far-field FoV of ±4 degrees at a maximum range of 250 m, and a near-field FoV of ±45 degrees at a range of 70 m. Its price decreases with volume, but is typically around a few thousand dollars and much more affordable than the Delphi SRR2 and ESR 2.5 radars used in CARMA Kit. We only use one radar since CARMA3 does not utilize radar data yet.

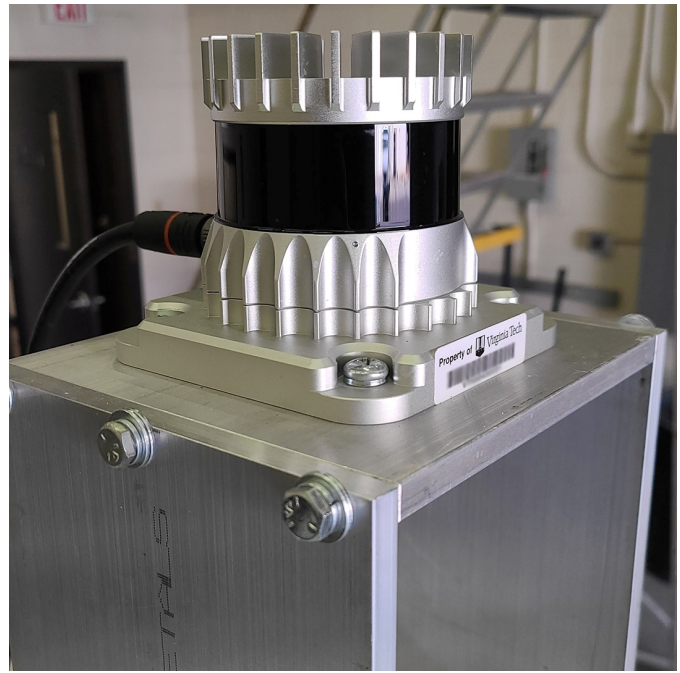

Fig. 6. Ouster OS1-64U lidar mounted on a level platform.

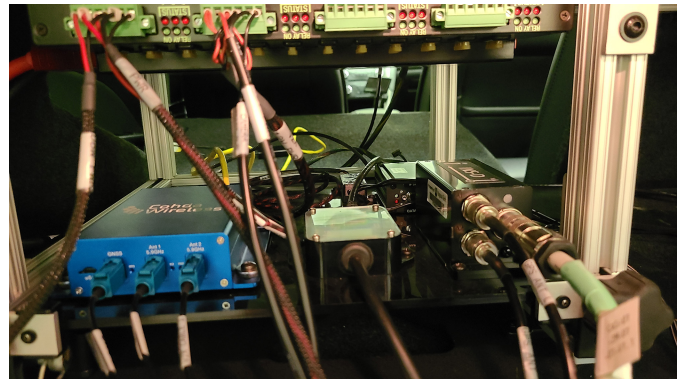

Fig. 7. Some of the equipment installed in the car's trunk. The iPDS can be seen at the top. Below it, the Inertial Labs INS-D module, Ouster Interface Box, and Cohda Wireless MK5 OBU are positioned from right to left. The Dataspeed CAN Gateway module is located behind the INS-D module.

The radar setup consists of the components listed in Table IV. The radar can be installed either behind the vehicle's front grills or on a bar attached to the front bumper. It should be connected to a CAN breakout board using the provided cable and connector kit. Further information can be found in [67], [68].

The Continental Radar Driver for CARMA GitHub repository contains the X-CAR driver developed for Continental ARS 408-21 radars [69]. Instructions are provided for downloading the driver code, building a Docker image of the driver, and setting up that image to launch with CARMA3. In addition to installation instructions, that repository also

## TABLE IV
### Radar Setup Components

| Quantity | Item |
|----------|------|
| 1 | Continental ARS 408-21 LRR |
| 1 | Continental adapter cable to CAN |
| 1 | Continental 8-pole connector kit |

## TABLE V
### GNSS/INS Setup Components

| Quantity | Item |
|----------|------|
| 1 | Inertial Labs INS-D module |
| 1 | Inertial Labs development board |
| 3 | RS232/RS422 to USB cable |
| 1 | Power adapter |
| 2 | Inertial Labs GNSS antenna |
| 2 | Antenna to INS-D TNC cable |

contains information on the driver's ROS API. The driver is forked from the ROS driver developed by the SZEnergy team with two modifications [70]. First, the X-CAR driver broadcasts a discovery message indicating its current status (operational, degraded, fault, or off) to CARMA3. Second, it publishes radar data in the format expected by CARMA3.

### D. GNSS/INS

A GNSS/INS module provides accurate position (latitude, longitude, and heading), velocity, and time information. It does so by receiving data from one or multiple global navigation satellite systems (GNSS) such as the American GPS (Global Positioning System), the Russian GLONASS (Global Navigation Satellite System), the Chinese BDS (BeiDou Navigation Satellite System), or the European Galileo, and fuses that information with data from its motion (accelerometer) and rotation (gyroscope) sensors that comprise an inertial navigation system (INS).

After careful evaluation of the products available on the market, we selected the Inertial Labs INS-D Dual Antenna GNSS/INS module [71]. It provides position, velocity, and time information with great precision, is IP67 and MIL-STD-810G certified, and is more affordable at about a third of the price of a NovAtel PwrPak7D-E2 used in the CARMA Kit.

The GNSS/INS setup consists of the components listed in Table V. The INS-D module should be mounted on a level surface in the trunk of the car, as shown in Fig. 7. In our setup the $x$ axis displayed on the module points to the right side of the car and the $y$ axis points directly forward (i.e. the connectors point directly to the back of the car).

The primary and secondary antennas should be mounted on top of the car with as much distance between them as possible, and connected to the INS-D module using the provided TNC cables. The INS-D module has to be connected to the development board, which itself should be connected to a power supply and the processing unit via a RS232/RS422 to USB cable.

The INS GUI program provided by Inertial Labs lets users configure their device to match their physical setup. The program allows users to select the output data format (we recommend selecting OPVT2A; or Orientation, Position, Velocity, and Time, 2 Antennas) and set the proper alignment of the INS-D module, the two antennas relative to the module, and the measuring point relative to the module. For X-CAR, the measuring point is the location of `base_link`, which CARMA3 considers to be the midpoint of the projection of the rear axle on the ground. The INS sensors (accelerometer, gyroscope, and magnetometer) are pre-calibrated and do not need calibration. Further information can be found in [72]–[74].

The Inertial Labs GNSS/INS Driver for CARMA GitHub repository contains the X-CAR driver developed for Inertial Labs GNSS/INS modules [75]. Instructions are provided for downloading the driver code, building a Docker image of the driver, and setting up that image to launch with CARMA3. Along with installation instructions, the repository also contains information on the driver's ROS API. The driver is forked from Inertial Labs' ROS driver with two modifications [76]. First, the X-CAR driver broadcasts a discovery message indicating its current status (operational, degraded, fault, or off). Second, instead of publishing the default Inertial Labs messages, it publishes fused GNSS/INS data as `gps_common/GPSFix` messages. It also publishes unfused INS data as `sensor_msgs/Imu` messages for cases where standalone INS data is needed.

### E. Power Distribution System

A power distribution system connects to the vehicle's 12V/24V battery and provides power to various devices installed in the vehicle. It also lets the user monitor the power consumption of different components.

Because Dataspeed provided our vehicle's DbW kit, we also used their intelligent Power Distribution System (iPDS) along with a Samlex America PST-1000-12 pure sine wave inverter [77]. The iPDS connects to the vehicle's 12V battery and provides 12 channels at 12V, up to 15A each, but the channels can also run at 24V, up to 10A if the car has a 24V battery. It is relatively affordable and offers programmable startup and shutdown functionality; over-current feedback and diagnostics; and CAN, Ethernet, and USB communication. The inverter connects to the car's 12V battery and provides 120V AC outlets that can be used to power the processing unit and other components as needed. Moreover, it can be controlled through the iPDS interface via an Ethernet connection.

The power distribution system consists of the components listed in Table VI. iPDS components are installed by Dataspeed. The iPDS module is installed in the trunk, the switch panel is installed in the storage bin between the front seats, and the touchscreen is installed in the front cup holder. The inverter should be safely installed in the vehicle's trunk and grounded. The iPDS does not require any calibration, but it can be accessed for configuration or diagnostics if needed. Further information can be found in [78].

TABLE VI
POWER DISTRIBUTION SYSTEM COMPONENTS

| Quantity | Item |
|----------|------|
| 1 | Dataspeed iPDS module |
| 1 | Dataspeed iPDS touchscreen |
| 1 | Dataspeed iPDS switch panel |
| 1 | Samlex America PST-1000-12 inverter |
| 1 | Samlex America DC-2000-KIT cable kit |

TABLE VII
DBW KIT COMPONENTS

| Quantity | Item |
|----------|------|
| 1 | Dataspeed Brake Throttle Emulator module |
| 1 | Dataspeed Shifting Interface module |
| 1 | Dataspeed Steering Interface module |
| 1 | Dataspeed CAN Gateway module |
| 1 | Dataspeed CAN USB breakout board |

TABLE VIII
V2X SETUP COMPONENTS.

| Quantity | Item |
|----------|------|
| 1 | Cohda Wireless MK5 OBU |
| 1 | Cohda Wireless DSRC/GNSS antenna |
| 1 | SD card |
| 1 | RJ45 cable |

## IV. ACTUATION

A DbW kit acts as an interface for communicating with the vehicle. Using a DbW kit, CARMA3 can receive vital information about the status of various vehicle functions (such as braking, steering, etc.) and send commands for controlling the vehicle. The DbW kit connects to the vehicle's CAN bus to read and publish messages. CAN bus message information (message IDs and their data format) are proprietary, so it takes an enormous effort to reverse-engineer the entire CAN bus, read CAN messages, and publish messages that bypass the built-in security features and fool the vehicle into thinking they are coming from a human driver. Therefore, most research institutions rely on a few companies that have done the reverse engineering and developed a DbW kit.

Dataspeed is the only company that offers a DbW kit for some Ford vehicles, including our 2017 Ford Fusion SE Hybrid. Therefore, we used the Dataspeed Drive-by-Wire Kit for X-CAR [79]. The DbW kit is a complete hardware and software system that enables electronic control of the car's brake, throttle, steering, and shifting, while maintaining safe operation by allowing a safety driver to override the system at any moment. The DbW kit also includes Dataspeed's Universal Lat/Lon Controller (ULC) that receives desired speed and angular velocity commands and publishes the appropriate steering, braking, and throttle commands to the car's CAN bus. This function removes the need for additional software such as AutonomouStuff's Speed and Steering Control (SSC) that is used in the CARMA Kit [80]. The DbW kit is by far the most expensive part of X-CAR with a mid-five-figure price tag.

The DbW kit consists of the components listed in Table VII. DbW kit components are installed by Dataspeed and the user should just connect the CAN Gateway module to the processing unit using a USB cable. The DbW kit does not require any calibration, but internal parameters can be changed and the firmware can be updated if needed. For further information consult [81].

The Dataspeed Drive-by-Wire Kit CAN Driver for CARMA and Dataspeed Universal Lat/Lon Controller (ULC) Driver for CARMA GitHub repositories contain the X-CAR drivers developed for the DbW kit and ULC [82], [83]. Instructions are provided for downloading driver codes, building Docker images of the drivers, and setting up those images to launch with CARMA3. Along with installation instructions, the repositories contain information about the drivers' ROS API. The drivers are forked from Dataspeed's ROS drivers with two modifications [84], [85]. First, they each broadcast a discovery message indicating their current status (operational, degraded, fault, or off). Second, the CAN driver publishes critical vehicle information reported by the CAN bus in the message formats required by CARMA3, while the ULC driver is configured to receive an Autoware Vehicle Command message for controlling the car.

## V. V2X

DSRC is a wireless communication technology that enables vehicles to directly communicate with one another and other road users, without relying on cellular equipment or other infrastructure. These one-way or two-way communication channels are specifically designed for automotive use along with a corresponding set of protocols and standards.

Similar to CARMA Kit, we use Cohda Wireless' MK5 On-Board Unit (OBU) which is a small, rugged module for DSRC. It features an antenna packaged with dual IEEE 802.11p wireless antennas along with a GNSS antenna. With a price tag of a few thousand dollars, it is reasonably affordable. We plan on adding a 5G-capable cellular router as well, most likely Cradlepoints's R1900 Series 5G Ruggedized Router [86].

The DSRC setup consists of the components listed in Table VIII. The OBU should be installed in the vehicle's trunk and connected to a 12V power supply, as shown in Fig. 7. It should be connected to the antenna installed on the car's roof via the three cables attached to the antenna (two for wireless data and one for GNSS data). Finally, it should be connected to the network switch (or the processing unit) via an RJ45 cable.

## VI. COMPUTATION AND DATA PROCESSING

The processing unit is installed on a vibration-absorbant surface in the vehicle's trunk and handles the computational load of CARMA3. We initially considered commercial options such as Crystal Rugged AVC0161, AutonomouStuff Spectra 2, and Nvidia Drive AGX Developer Kit [87]–[89]. However, because of their mid-five-figure price tag, as well as some's lack of support for newer components (e.g. only supporting older 8th and 9th generation Intel Core and Xeon processors and not supporting AMD products), we decided to build our own processing unit using available consumer parts. Since some of the components (notably the case, motherboard, and RAM) are not ruggedized, our choice increases the risk of vibration damage. However, given that our processing unit is installed on a vibration-absorbing surface and we do not plan on driving on rough terrain, we think the risk of vibration damage is low. Table IX lists processing unit components. The total cost of the unit was less than $3,500 and with about 10 TFLOPS of FP32 performance it is comparable to an Nvidia Drive AGX Pegasus.

TABLE IX
PROCESSING UNIT COMPONENTS

| Item | Component |
|------|-----------|
| Case | SilverStone GD08 |
| CPU | AMD Ryzen 9 5900X |
| Motherboard | Gigabyte X570 AORUS Master ATX |
| CPU Cooler | Noctua NH-C14S |
| RAM | G.Skill Trident Z RGB 64 GB (2×32 GB) DDR4-3600 CL18 |
| Graphics Card | EVGA Nvidia GeForce RTX 3060 XC Gaming |
| Power Supply Unit | EVGA SuperNOVA P2 750W 80+ Platinum Fully Modular ATX |
| Boot Drive | Samsung 980 Pro 1 TB M.2-2280 NVMe SSD |
| High-Speed Storage Drives | 2×Samsung 980 Pro 2 TB M.2-2280 NVMe SSD |
| Network Interface Card | Intel X710-DA2 Dual-Port SFP+ 10GbE |
| Case Fan | 5×Arctic P12 PST 120 mm |
| Small Case Fan | 2×Arctic F8 80 mm |

The processing unit runs Ubuntu 20.04.4 LTS, which is installed on the boot drive. The two high-speed storage drives are combined in a mirrored ZFS pool, which is equivalent to a RAID 1 configuration. The pool shows up as a single volume and data written to this volume is written to both drives so that if one fails, data is kept safe on the other one.

## VII. CARMA3

CARMA3 is a reusable, extensible platform for controlling CAVs. It provides a rich, generic API for a variety of sensing and control hardware, as well as for third-party plugins that implement guidance algorithms to plan vehicle trajectories. Several Autoware® modules are used in CARMA3 either directly or with some modifications. An overview of CARMA3 can be seen in Fig. 8.

AV motion generally involves four aspects. The first is localization, which is the act of determining the vehicle's current location with respect to the earth and with respect to its planned route (the desired path of travel). This is handled by the Localization subsystem of CARMA3. The second is perception, which includes detecting and classifying objects, lane markings, and traffic signs and signals, tracking the motion of moving objects, and predicting their future path. These tasks are performed by the World Model subsystem of CARMA3.

The third aspect is guidance, also known as trajectory planning, which is the process of determining how the vehicle is to move from its current location to its destination. The destination and route will be passed on to the guidance algorithms, and they then decide how the vehicle's motion needs to be adjusted at any time in order to follow the route. This aspect is handled by the Guidance Core subsystem of CARMA3. Alongside it, the Guidance Plugins subsystem handles strategic, tactical, and control plugins.

The final aspect of AV motion is control, which covers the actuation of the vehicle's physical devices to induce changes in motion (for cars these typically include causing the wheels to rotate faster or slower and turning the steering wheel). Therefore, Localization and World Model outputs become inputs to Guidance Core, and the output of Guidance Core becomes an input to the control function. Finally, the V2X subsystem of CARMA3 is responsible for converting a variety of V2X messaging protocols that enable cooperative perception and navigation, including tasks such as platooning and cooperative intersection management.

To provide more flexibility, each CARMA3 repository is distributable as a Docker image containing the compiled applications and their dependencies. These images will be built upon a base image called `carma-base` which contains (and abstracts away) the setup of the ROS system, user configuration, and installation of any dependencies of CARMA3. Each hardware driver developed for X-CAR is distributable as a Docker image as well. Refer to [10], [90], [91] for further information. A guide to installing CARMA3 on X-CAR, configuring the software, and operating X-CAR is provided in Appendix A.

## VIII. MAPS AND ROUTES

The Localization subsystem of CARMA3 is responsible for determining and reporting the most accurate estimate of the vehicle's position, velocity, acceleration, and heading at each instant. This is done by using information from the lidar, the GNSS/INS module, a point cloud map, and a vector map.

CARMA3 uses a number of coordinate frames. The *earth (world)* coordinate frame is an Earth-centered, Earth-fixed (ECEF) coordinate system with its origin at the Earth's center of gravity where each location is determined by latitude and longitude coordinates. The *map* coordinate frame is an

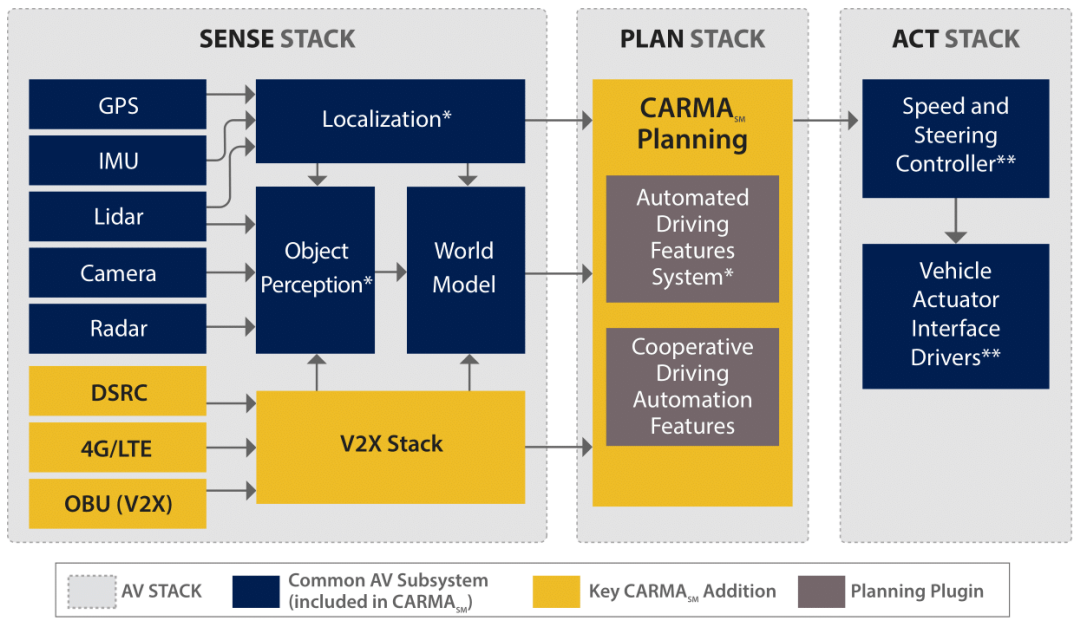

**AV** - Automated Vehicle, **CARMA** - Cooperative Automation Research Mobility Applications, **GPS** - Global Positioning System, **IMU** - Inertial Measurement Unit, **OBU** - On-Board Units, **V2X** - Vehicle-to-Everything
* Supported by Autoware, **Supported vehicle controllers: Dataspeed, PACMOD, and New Eagle.

Fig. 8. Overview of CARMA3 [10].

ENU (East North Up) coordinate frame with its origin at the starting point of the point cloud map, which is read from the `geoReference` tag of the vector map. Finally, the *base_link* coordinate frame is a FLU (Forward Left Up) coordinate frame attached to the vehicle's body, with its origin at `base_link`. Each sensor has its own coordinate frame, though the origin and axis orientation vary from sensor to sensor. These coordinate frames are defined in a ROS Universal Robot Description Format (URDF) file.

### A. Point Cloud Map

A point cloud map is comprised of data points in space that represent the three-dimensional shape of the surrounding environment. To generate a point cloud map of an environment for autonomous driving, a car needs to drive around and scan that environment with its lidar sensor(s).

Individual lidar scans (each lidar scan is a point cloud generated at a specific instant) are usually merged to form a map using a process called NDT mapping [92]. In NDT mapping, each lidar scan is transformed into a grid-based representation, where each grid cell is approximated with a multivariate Gaussian distribution. Along with knowledge of the vehicle's motion (usually obtained from an IMU), this representation enables finding a transformation between consecutive scans, filtering out moving objects, and allowing stationary areas to be merged together to generate a map. More information on this topic is available at [93].

### B. Vector Map

A vector map is a vector representation of map data and traffic rules. CARMA3 employs the Lanelet2 vector map framework. It is based on *Lanelets*, or atomic lane sections that combine to form the road network through their neighborhood relationship. Traffic rules are defined by *Regulatory Elements*, which represent objects like traffic lights and stop signs [94].

A Lanelet2 map includes a Physical Layer, which contains the observable (usually real) elements and a Relational Layer, where the elements of the Physical Layer are connected to lanes, areas, and traffic rules. The Topological Layer is defined implicitly from the context and neighborhood relationships of the Relational Layer, where the elements of the Relational Layer are combined into a network of potentially passable regions that depend on the driving scenario and road user [94].

A Lanelet2 map consists of five elements: Points and Line Strings belonging to the Physical Layer and Lanelets, Areas, and Regulatory Elements that belong to the Relational Layer. Each element is identified by a unique ID (which is useful for an efficient construction of the Topological Layer) and can be assigned attributes in the form of key-value pairs. Some of these attributes are fixed, but others can be used to improve the map [94].

We used RoadRunner, which is a proprietary software, for vector map generation [95]. However, free and open-source alternatives such as Tier IV's Vector Map Builder are also available [96].

### C. Routes

A route in CARMA3 is a set of one or multiple destination points that the vehicle has to reach consecutively. A CARMA3 route file is a `.csv` file that has four columns, representing latitude, longitude, ID, and description of each destination point, respectively. When creating a route file, the user needs to ensure that a viable path connecting consecutive destination points exists on the vector map.

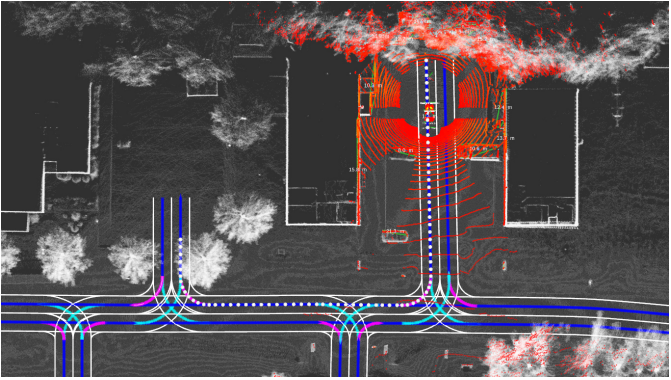

Fig. 9. Autonomous X-CAR Operation.

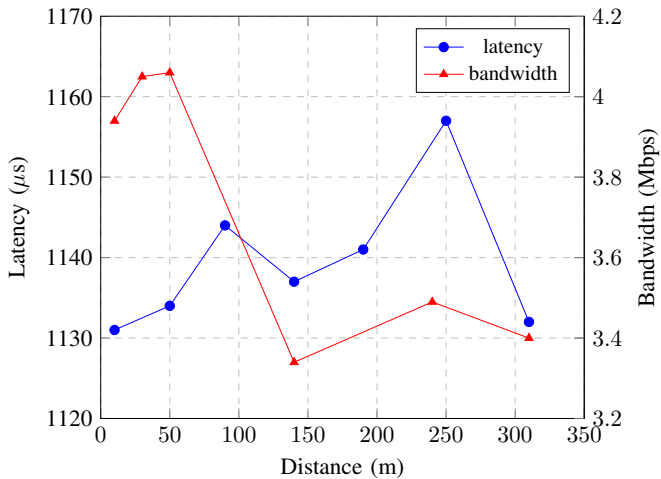

Fig. 10. Communication latency and bandwidth with respect to distance between two identical Cohda Wireless MK5 OBUs.

## IX. X-CAR Performance

After installing CARMA3 on the processing unit, configuring it to work with the X-CAR sensor suite, and launching it alongside X-CAR drivers, we experimented with X-CAR in a limited capacity since full testing of cooperative behavior requires at least two X-CARs and we only had one. One experiment involved testing the autonomy aspects of X-CAR. As Fig. 9 shows, X-CAR can localize itself on the map, detect surrounding objects and their distance, plan a trajectory, and execute that trajectory to reach its destination. Another experiment involved evaluation of the latency and bandwidth of communication between two identical MK5 OBUs along a straight, open road. The results, depicted in Fig. 10, show that communication latency is virtually constant and around 1.1 ms, which is expected. They also show that at the default Modulation & Coding Scheme (MCS) of R12QPSK, communication bandwidth is an expected 4 Mbps at small distances, but drops to around 3.4 Mbps at larger distances which can be attributed to increasing packet loss. Based on Fig. 10, we estimate the unobstructed communication range at the default MCS to be around 350 m.

## X. Conclusions

This paper introduced X-CAR, an experimental platform for connected autonomy research powered by CARMA3. After providing the necessary background, we discussed the communication and sensing suite employed to produce a comprehensive understanding of the vehicle's surrounding environment. We also discussed vehicle actuation, as well as the hardware required for computation and data processing. Finally, we discussed maps and routes as well as X-CAR's real-world performance. Overall, X-CAR aims to increase the versatility of CARMA3 and facilitate its adoption for R&D of CDA, with future work focusing on its autonomous and cooperative aspects.

## Acknowledgment

The authors wish to express their gratitude to the people who assisted in the development of X-CAR, including Professor Al Wicks, Professor Steve Southward, Xihui Wu, Bhavi Bharat Kotha, Dataspeed and especially Chris Capaldi who helped with the installation of equipment our, as well as the CARMA Platform$^{\text{SM}}$ development team and in particular Michael McConnell and Safak Ercisli. The authors also would like to thank the support teams at Ouster and Inertial Labs.

## Appendix A
## A Guide to CARMA3

In what follows, Appendix A-A provides a guide to installing CARMA3 on X-CAR, Appendix A-B discusses CARMA3's configuration and calibration files, and Appendix A-C provides a step-by-step guide to operating X-CAR. Further information can be found in the X-CAR Reference Manual [11].

### A. Installing CARMA3 on X-CAR

The CARMA3 development team has provided excellent instructions for setting up CARMA3 on their vehicles in [97]. As setting up CARMA3 on X-CAR is very similar, we ask the reader to follow those instructions and only provide what the reader should do differently below.

1) Skip the Setup VMWare Image for Development page since CARMA3 will be installed on a native Ubuntu 20.04 machine.
2) Ensure the latest Nvidia driver is installed. The driver and CUDA version can be checked by running the `nvidia-smi` shell command.
3) CUDA 11.2 step: since CUDA has already been installed when installing the Nvidia driver, skip the `# install cuda` portion of the code.
4) Cloning The Repository step: instead of line 3, run the following shell command.

```
1  curl -L https://raw.githubusercontent.com/VT-A
   ↪ SIM-LAB/carma-platform/develop/carma-platf
   ↪ orm.repos | vcs
   ↪ import
```

This ensures that X-CAR's drivers, configuration files, and repositories are cloned instead of the standard CARMA3 ones.

5) Setup The `/opt/carma` Folder step: instead of line 2, run the following shell command.

```
1  sudo bash ~/opt_carma_setup.bash
   ↪ ~/carma_ws/src/carma-config/example_calibr
   ↪ ation_folder/vehicle/
```

6) Stop after finishing the Setup The `/opt/carma` Folder step. If you proceed, the instructions will download the standard CARMA3 Docker images. Since we have made some modifications to the code, our Docker images have to be built from source.

7) The `carma-platform` Docker image is a child of the `autoware.ai` Docker image, which itself is a child of the `carma-base` Docker image. Therefore, to build these Docker images from source you have to first build `carma-base`, then `autoware.ai`, and finally `carma-platform`. You also have to build the `carma-web-ui` and `carma-msgs` Docker images, as well as that of any hardware driver you want to use. To build these images from source make sure that all local repositories are synced with their respective GitHub ones. Then, run the following shell commands. Building a Docker image can take several minutes, so wait until each build is finished before moving onto the next image.

```
1  cd ~/carma_ws/src/carma-base/docker
2  sudo ./build-image.sh -d
3  cd ../../autoware.ai/docker
4  sudo ./build-image.sh -d
5  cd ../../carma-platform/docker
6  sudo ./build-image.sh -d
7  cd ../../carma-msgs/docker
8  sudo ./build-image.sh -d
9  cd ../../carma-web-ui/docker
10 sudo ./build-image.sh -d
11 cd ../../carma-cohda-dsrc-driver/docker
12 sudo ./build-image.sh -d
13 cd ../../lucid_camera_driver/docker
14 sudo ./build-image.sh -d
15 cd ../../ouster_lidar_driver/docker
16 sudo ./build-image.sh -d
17 cd ../../continental_radar_driver/docker
18 sudo ./build-image.sh -d
19 cd ../../inertiallabs_gnss_driver/docker
20 sudo ./build-image.sh -d
21 cd ../../dataspeed_can_driver/docker
22 sudo ./build-image.sh -d
23 cd ../../dataspeed_controller_driver/docker
24 sudo ./build-image.sh -d
```

### B. Configuring CARMA3

Core CARMA3 subsystems are designed to be agnostic of the platform (vehicle and sensor suite) they are installed on, which allows CARMA3 to be versatile and run on any vehicle and sensor suite that satisfies its API requirements. As a result, the configuration and calibration files required for operation of CARMA3 on different vehicles are stored in the ~/carma_ws/src/carma-config directory. A configuration folder in this directory contains vehicle configuration data that is specific to a particular class of vehicle and sensor suite, such as the launch files for sensor drivers or the Docker Compose files that launch CARMA3 with the proper configuration. In contrast, a calibration folder contains information that is specific to individual vehicles, such as precise sensor orientations or controller tuning.

X-CAR configuration files are located in the ~/carma_ws/src/carma-config/ford_fusion_2017 directory and include the following files, similar to the `ford_fusion_sehybrid_2019` folder of the standard CARMA3 installation.

- `bridge.yml`: contains explicit topic mappings between ROS 1 and ROS 2 portions of CARMA3 while ROS 2 migration is underway.
- `build-image.sh`: generic build script for CARMA3 configuration Docker images.
- `carma_docker.launch`: the ROS launch file for the CARMA3 ROS network. Launches the required ROS nodes and sets up the parameter server. Also sets up all static transforms used by TF2 within the system. Sets the path for vehicle configuration and calibration directories, route files, and point cloud and vector maps.
- `carma_rosconsole.conf`: sets the default ROS output level as well as package-specific log levels. It can be useful for debugging issues with specific packages.
- `docker-compose-background.yml`: launches background Docker containers required for running CARMA3.
- `docker-compose.yml`: launches Docker containers required for running CARMA3. Alongside launching CARMA3, it allows the user to launch the desired mock and real drivers.
- `drivers.launch`: launches the ROS nodes of desired mock or real drivers.
- `VehicleConfigParams.yaml`: provides the ROS parameters that define the characteristics of the host vehicle configuration.

X-CAR calibration files are located in the ~/carma_ws/src/carma-config/ford_fusion_2017_calibration_folder/vehicle/calibration directory and include the following files, similar to the `example_calibration_folder/vehicle/calibration` folder of the standard CARMA3 installation.

- `identifiers/UniqueVehicleParams.yaml`: provides the ROS parameters that define a unique vehicle's specific traits.
- `lidar_localizer/ndt_matching/params.yaml`: defines TF2 transformation parameters used for NDT mapping and matching.
- `pure_pursuit/calibration.yaml`: pure pursuit look-ahead calibration parameters.
- `range_vision_fusion`: contains lidar-camera calibration files.
- `urdf/carma.urdf`: defines static TF2 transformations that provide the position of different hardware components. All transformations are relative to `base_link`, and should be defined accurately to ensure proper operation.

## C. Operating X-CAR

Before operating X-CAR, ensure that all necessary Docker images for CARMA3 and the desired drivers have been built. You can verify this by running the `docker images` shell command. Note that CARMA3 is still in development, so before operating X-CAR ensure that the vehicle is in an open area away from other cars and vulnerable road users. Have a safety driver behind the steering wheel and verify that the emergency stop button works.

Follow the steps below to operate X-CAR:

1) Run the following shell commands to build the latest configuration Docker image and set it as the configuration used by CARMA3.

```
1  cd ~/carma_ws/src/carma-config/ford_fusion_2017
2  sudo ./build-image.sh -d
3  carma config set usdotfhwastoldev/carma-config⌋
   ↪ :develop-ford-fusion-2017
```

2) CARMA3 requires CAN, controller, lidar, GNSS/INS, and camera drivers to run (otherwise it shuts down), though some of these can be mock drivers. At a minimum, CARMA3 needs the real CAN, controller, and lidar drivers to be functional. To start CARMA3, run the following shell command.

```
1  carma start all roscore platform platform_ros2
   ↪ ros1_bridge [can driver] [controller
   ↪ driver] [lidar driver] [gnss driver]
   ↪ [camera driver]
```

In the command above, each driver can be either a mock or real driver, but not both. For example, if you want to run X-CAR without cameras, run the following shell command to start CARMA3.

```
1  carma start all roscore platform platform_ros2
   ↪ ros1_bridge dataspeed-can-driver
   ↪ dataspeed-controller-driver
   ↪ ouster-lidar-driver
   ↪ inertiallabs-gnss-driver mock-camera-driver
```

3) Copy the `carma_default.rviz` file from the ~/carma_ws/src/carma-platform/carma/rviz directory into the `/opt/carma/maps` directory. Open a new Terminal tab and run the following shell command to launch RViz.

```
1  carma exec rviz -d
   ↪ /opt/carma/maps/carma_default.rviz
```

4) Verify that the sensors are working properly. You can visualize lidar and camera data inside RViz. To check the output of the GNSS/INS module and the Dataspeed DbW kit, open a new Terminal tab and run the `carma exec` shell command to open a shell inside the usdotfhwastoldev/carma-platform:develop Docker image. Then use `rostopic echo` to view the output of the relevant ROS topics.

5) Wait for the vector and point cloud maps to load. When they are loaded they will be shown in RViz.

6) In Google Chrome (or Chromium), go to `localhost` or `127.0.0.1`. Click on the **Start Platform** button. You can turn on **Debug Mode** from the lower right corner of the screen.

7) In the new screen, click on **Logs** and verify that *All essential drivers are ready* message is displayed.

8) On the top right corner, verify that the localization status indicator is green. You can also verify accurate localization in RViz, where the real-time lidar scan (shown in red) should match the point cloud map (shown in white).

9) Select a route from the list of routes. If the route is valid, RViz will show the generated waypoints from the vehicle to the final destination, similar to Fig. 9.

10) In Google Chrome, click on **Status** and check the **Robot Active** and **Robot Enabled** fields. Both will likely report **false**.

11) Open a new Terminal tab and run the `carma exec -i usdotfhwastoldev/carma-dataspeed-controller-driver:develop` shell command to open a shell inside the `usdotfhwastoldev/carma-dataspeed-controller-driver:develop` Docker image. Then run the following shell commands.

```
1  . ./devel/setup.bash
2  rostopic pub /vehicle/dbw_enabled std_msgs/Bool
   ↪ '{data: True}'
```

Check Google Chrome and verify that **Robot Active** now reports **True**. Then terminate `rostopic pub`, but do not exit the Docker image shell.

12) In the same Docker image shell, run the following command.

```
1  rostopic pub /ulc_cmd dataspeed_ulc_msgs/UlcCmd
   ↪ '{enable_pedals: True, enable_shifting:
   ↪ True, enable_steering: True,
   ↪ shift_from_park: True}'
```

This will signal to the ULC to be ready to receive control commands. You will likely hear a siren from the car. Press on the brakes and then terminate the command. This will set the `override` flag of the ULC to 1, which we will clear in Step 14.

13) In Google Chrome, click on the big green button on the lower left side. It should start glowing.

14) In the same Docker image shell as above, run the following command.

```
1  rostopic pub /ulc_cmd dataspeed_ulc_msgs/UlcCmd
   ↪ '{clear: True}'
```

The vehicle should start driving and the big glowing green button in Google Chrome should turn blue. The vehicle will drive towards designated destinations and stop after reaching the last destination.

15) Once you are finished, terminate CARMA3, RViz, and exit any shell environment created inside a Docker image. Then, run the following shell command.

```
1  carma stop all
```

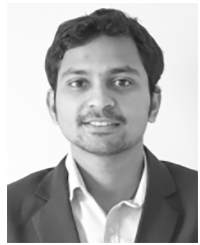

**Anshul Nayak** received his B.Sc. degree in mechanical engineering from NIT, Rourkela, India in 2015. He completed his Master's degree in Mechanical engineering in 2020 at Virginia Tech and is currently pursuing his Ph.D degree at the same university. His research interests include cooperative planning and uncertainty estimation in prediction.

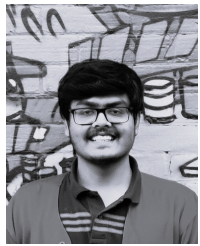

**Darshit Patel** received his B. Tech. degree in mechanical engineering from NIT Surat, India in 2020 and is currently pursuing a Master's degree in mechanical engineering at Virginia Tech, USA. His research interests include motion planning and control of autonomous and connected autonomous vehicles.

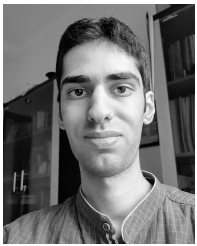

**Goodarz Mehr** received his B.Sc. degree in mechanical engineering from Sharif University of Technology, Tehran, Iran, and is currently pursuing a Ph.D. degree in mechanical engineering from Virginia Tech. His research interests include robotics and control, stochastic planning models, machine learning, and cooperative perception.

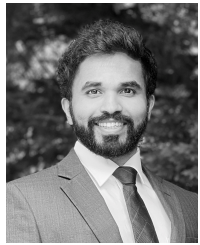

**Shathushan Sivashangaran** received his B.S. degree (Summa Cum Laude) and M.S. degree, both in mechanical engineering, from the State University of New York at Buffalo, USA. He is currently pursuing a Ph.D. degree in mechanical engineering from Virginia Tech at the Autonomous Systems and Intelligent Machines (ASIM) Laboratory. His research interests include robotics, deep reinforcement learning, intelligent autonomous navigation, and adaptive control.

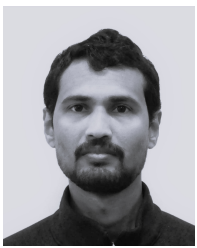

**Prasenjit Ghorai** received his B.Tech. degree from Maulana Abul Kalam Azad University of Technology (formerly West Bengal University of Technology, India) in electronics and instrumentation engineering, his M.Tech. degree in control & instrumentation engineering from the University of Calcutta, and his Ph.D. degree in engineering from the National Institute of Technology (NIT) Agartala (in collaboration with Indian Institute of Technology, Guwahati, India). He was an Assistant Professor of Electronics and Instrumentation Engineering with NIT Agartala from 2011 to 2019. He is currently working as a Postdoctoral Associate at the Autonomous Systems and Intelligent Machines Laboratory at Virginia Tech and conducts research on cooperative and connected autonomous vehicles.

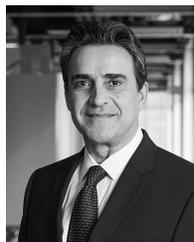

**Azim Eskandarian** has been a Professor and Head of the Mechanical Engineering Department at Virginia Tech since August 2015. He became the Nicholas and Rebecca Des Champs chaired Professor in April 2018. He also has a courtesy appointment as a Professor in the Electrical and Computer Engineering Department since 2021. He established the Autonomous Systems and Intelligent Machines laboratory at Virginia Tech and has conducted pioneering research in autonomous vehicles, human/driver cognition and vehicle interface, advanced driver assistance systems, and robotics. Before joining Virginia Tech, he was a Professor of Engineering and Applied Science at George Washington University (GWU) and the Founding Director of the Center for Intelligent Systems Research, from 1996 to 2015, the Director of the Transportation Safety and Security University Area of Excellence, from 2002 to 2015, and the Co-Founder of the National Crash Analysis Center in 1992 and its Director from 1998 to 2002 and 2013 to 2015. From 1989 to 1992, he was an Assistant Professor at Pennsylvania State University, York, PA, and an Engineer/Project Manager in the industry from 1983 to 1989. Dr. Eskandarian is a Fellow of ASME, a member of SAE, and a Senior Member of IEEE professional societies. He received SAE's Vincent Bendix Automotive Electronics Engineering Award in 2021, IEEE ITS Society's Outstanding Researcher Award in 2017, and GWU's School of Engineering Outstanding Researcher Award in 2013.

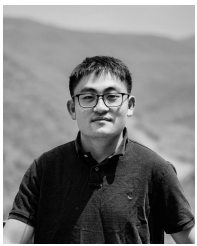

**Ce Zhang** received the B.Sc. degree in mechanical engineering from Virginia Tech (VT), USA in 2018. He is currently pursuing a Ph.D. degree in mechanical engineering at the Virginia Tech Autonomous System and Intelligent Machine (ASIM) Lab. His research interests include automated driving, driver cognitive study, human machine interface, and machine learning.